\PassOptionsToPackage{prologue,dvipsnames}{xcolor}
\documentclass[sigconf]{acmart}

\usepackage{graphicx}
\usepackage{xspace}
\usepackage{amsmath}
\usepackage{booktabs} 
\usepackage{mathrsfs}
\usepackage[utf8]{inputenc}
\usepackage{microtype}
\usepackage{dsfont}
\usepackage{subfigure}
\usepackage{CJKutf8}
\usepackage{bigstrut}
\usepackage{multirow, makecell}
\usepackage[normalem]{ulem}
\usepackage{amsfonts}
\usepackage{multicol}
\usepackage{tipa}
\usepackage{bm}
\usepackage[ruled,linesnumbered]{algorithm2e}
\usepackage{paralist}
\usepackage{arydshln}
\usepackage{verbatim}
\usepackage{csquotes}
\usepackage{mdwlist}
\usepackage{makecell}
\usepackage{array}
\usepackage{colortbl}
\usepackage{tabularx}
\usepackage{bbm}
\usepackage{tikz}
\usepackage{bbding}
\usepackage{balance}
\usepackage{latexsym}
\usepackage{microtype}
\usepackage{enumerate}
\usepackage{enumitem}
\usepackage{pifont}
\usepackage{color} 
\usepackage{subfigure}
\usepackage{listings}
\usepackage{tcolorbox}
\tcbuselibrary{breakable}
\tcbuselibrary{skins}

\newcommand{\advhot}{\textsc{AdvHotpotQA}\xspace}
\newcommand{\wiki}{\textsc{2WikiMultiHopQA}\xspace}

\newcommand{\gsm}{\textsc{GSM8K}\xspace}
\newcommand{\mat}{\textsc{MATH}\xspace}

\newcommand{\cott}{{CoT}\xspace}
\newcommand{\llm}{{LLMs}\xspace}

\newcommand{\ve}{{VE}\xspace}

\newcommand{\ours}{\textsc{PSSD}\xspace}
\newcommand{\ourssft}{\textsc{PSSD-SFT}\xspace}
\newcommand{\cotori}{{CoT-Ori}\xspace}
\newcommand{\react}{{ReAct}\xspace}
\newcommand{\stand}{{Stand}\xspace}
\newcommand{\cotsc}{{CoT-SC}\xspace}

\newcommand{\tran}{{TRAN}\xspace}

\newcommand{\lema}{\textsc{LeMa}\xspace}
\newcommand{\salam}{\textsc{SALAM}\xspace}
\newcommand{\gwfs}{\textsc{GWfS}\xspace}
\newcommand{\srethink}{{Self-rethinking}\xspace}
\newcommand{\sta}{\textsc{STaR}\xspace}
\newcommand{\scontrast}{{Self-Contrast}\xspace}

\newcommand{\mi}{intuition-based id\xspace}
\newcommand{\mii}{rule-driven superego\xspace}
\newcommand{\miii}{script-centric ego\xspace}

\newcolumntype{L}[1]{>{\raggedright\let\newline\\\arraybackslash\hspace{0pt}}m{#1}}
\newcolumntype{C}[1]{>{\centering\let\newline\\\arraybackslash\hspace{0pt}}m{#1}}
\newcolumntype{R}[1]{>{\raggedleft\let\newline\\\arraybackslash\hspace{0pt}}m{#1}}

\AtBeginDocument{%
  }

\setcopyright{acmlicensed}
\copyrightyear{2025}
\acmYear{2025}
\acmDOI{10.1145/3696410.3714715}
\acmConference[WWW '25]{Proceedings of the ACM Web Conference 2025}{April 28-May 2, 2025}{Sydney, NSW, Australia}
\acmISBN{979-8-4007-1274-6/25/04}

\begin{document}

\title[\ours: Making LLMs Self-denial via Human Psyche Structure]{\ours: Making Large Language Models Self-denial \\ via Human Psyche Structure}

\author{Jinzhi Liao}
\orcid{0000-0002-2898-6559}
\affiliation{%
  \department{Laboratory for Big Data and Decision,}
  \institution{National University of Defense Technology}
  \city{Changsha}
  \state{Hunan}
  \country{China}
}
\email{liaojinzhi12@nudt.edu.cn}

\author{Zenghua Liao}
\orcid{0009-0002-3155-2353}
\affiliation{%
  \department{Laboratory for Big Data and Decision,}
  \institution{National University of Defense Technology}
  \city{Changsha}
  \state{Hunan}
  \country{China}
}
\email{liaozenghua18@nudt.edu.cn}

\author{Xiang Zhao}
\orcid{0000-0001-6339-0219}
\authornote{Corresponding author}
\affiliation{%
  \department{Laboratory for Big Data and Decision,}
  \institution{National University of Defense Technology}
  \city{Changsha}
  \state{Hunan}
  \country{China}
}
\email{xiangzhao@nudt.edu.cn}

\renewcommand{\shortauthors}{Jinzhi Liao, Zenghua Liao and Xiang Zhao}

\begin{abstract}
  The enhance of accuracy in reasoning results of LLMs arouses the community's interests, wherein pioneering studies investigate post-hoc strategies to rectify potential mistakes.
  Despite extensive efforts,
  they are all stuck in a state of resource competition demanding significant time and computing expenses.
  The cause of the situation lies in the failure of identifying the fundamental feature of the solutions in this line, coined as the self-denial of LLMs.
  In other words, LLMs should confidently determine the potential existence of mistakes and carefully execute the targeted correction.
  As the whole procedure conducts within LLMs, supporting and persuasive references are hard to acquire, while the absence of specific steps towards refining hidden mistakes persists even when errors are acknowledged.  
  In response to the challenges, we present \ours, which refers to and implements the human psyche structure such that three distinct and interconnected roles contribute to human reasoning.
  Specifically, \ours leverages the recent multi-agent paradigm, and is further enhanced with three innovatively conceived roles:
  (1) the \mi role that provides initial attempts based on benign LLMs;
  (2) the \mii role that summarizes rules to regulate the above attempts, and returns specific key points as guidance;
  and (3) the \miii role that absorbs all procedural information to generate executable script for the final answer prediction.
  Extensive experiments demonstrate that the proposed design not only better enhance reasoning capabilities, but also seamlessly integrate with current models,  leading to superior performance.
\end{abstract}




\maketitle

\vspace{-1ex}
\section{Introduction}
\label{sec:intro}

Recently, the NLP community has witnessed the booming development of \llm, where many downstream tasks manifested new milestones, especially in reasoning tasks~\cite{DBLP:conf/naacl/PatelBG21,DBLP:conf/nips/TalmorYBBGCB21}. 
Albeit superior, \llm still exhibit weakness in guaranteeing the reasoning correctness.
That is, \llm tend to make up facts and details, hence \textit{misleading} the direction of inference and generating the errorneous results~\cite{DBLP:conf/icml/ShiCMSDCSZ23,DBLP:conf/iclr/ZhouSHWS0SCBLC23,DBLP:journals/corr/abs-2307-15043}.

Amidst this backdrop, pioneering studies have investigated the post-hoc strategy, which emphasizes refining generated results, primarily influenced by the concept of \textit{correcting based on mistakes}~\cite{Ito1998NegativeIW}.
According to the achievement, these methods mainly can be put into three categories:
(1) \textit{Fine-tuning \llm} aims to proactively prevent the recurrence of mistakes by   
fine-tuning on previous correct and incorrect samples~\cite{DBLP:journals/corr/abs-2310-20689,DBLP:conf/acl/TongLWWTS24,DBLP:journals/corr/abs-2302-13007,DBLP:journals/corr/abs-2407-00497}; 
(2) \textit{Leveraging tools} iteratively polishes results through interaction with external assistance
to correct the mistakes in each step~\cite{DBLP:conf/nips/ShinnCGNY23,DBLP:conf/iclr/YaoZYDSN023,DBLP:conf/acl/ZhaoLJQB23,DBLP:conf/emnlp/YangLL23};
(3) \textit{Multi-agent debate} defines multiple roles of LLMs, with each individually generating a response and engaging in multi-round debates to reach a consistent answer~\cite{DBLP:journals/corr/abs-2305-14325,DBLP:conf/emnlp/CohenHGG23,DBLP:journals/corr/abs-2305-19118,DBLP:conf/acl/ZhangSWPWZ024}.
The training phase in first class does not meet the timely demands in practical applications.
Though methods in second class can provide immediate results, their performance relies heavily on availability of high-quality external resources. 
In order to ensure the timeliness and avoid the mentioned dependency, our research focuses on exploiting the \textit{inherent potential} of \llm, which belongs to last stream
(vividly shown in Figure~\ref{fig:comparison}).

As aforementioned, current studies in multi-agent debate try to allocate several roles for agents initializing discussions, where the round of interaction depends on reach of the consistency and number of roles relies on the empirical study.
These intuitive designs, instead of approaching the fundamental features of learning from mistakes, are stuck in the \textit{resource competition} (ref. Figure~\ref{tab: resource}), in which satisfactory results are highly dependent on the times of invoking LLMs.
To avoid the dilemma, we first identify the feature in this category as: \textit{facilitating agents' self-denial}, where the reasoning direction of LLMs is guided and refined via their experience.
These enhanced agents consciously exert control over the generation by appropriate interventions, corrections, or completions in reasoning, based on confidential determination.
The process theoretically prevents a stack of resources and obtains gold results.

The identification, meanwhile, reveals notorious challenges in pursuit of the goal.
(1) To ensure accurate denial, LLMs must maintain confidentiality in their determinations of results, which necessitates persuasive evidences. 
Without access to external resources, the acquisition of supporting information through direct means is limited.
(2) The rectification of the wrong generation after denial is even more challenging.
Reasons contributing to correct predictions are intricate and imperceptible, making LLMs fail to reach the correct answer from the outset.
(3) It is significant to control over required resources. 
When the procedure merely relies on countless refinements, it risks becoming a variant of resource competition.

To address the task, we refer to Freudian theory of human psyche~\cite{freud1923das}, in which \textit{the id, the ego}, and \textit{the superego} coexist simultaneously to attempt, regulate and adapt human behaviors; that is, human undergoes growth through internal debating among three personalities.
The theory is in line with our goal to acquire the intuitive attempt, regular guidance, and detailed rectification via multi-agent framework.
In this connection, we are motivated to implement the human psyche structure, and conceive a \underline{p}syche \underline{s}tructure for \underline{s}elf-\underline{d}enial (i.e., \ours) of \llm, and further enhanced by exploiting tailored designs of multiple agents, thereby comprehensively activating \llm' inherent capabilities.

Specifically, we design an \textit{\mi} role that functions as an innate drive, solely utilizing the \llm' reasoning ability to answer the given question. 
Through directly generating multiple reasoning paths, the original attempts are collected as intuitive responses.
Second, another LLM performs as the \textit{\mii} role, embodying a rational entity which provides supporting references from methodological perspective.
The rules, derived and processed empirically from training data, guide the role in abstracting key points of reasoning to support the denial.
Third, the \textit{\miii} role serves as a mediator to bridge the aforementioned roles.
This agent generates an executable script that adheres to intuitive attempts and key points, guiding its detailed execution to produce the final answer.
Last but not least, as the novel cooperativity of roles in \ours, different from other multi-agent methods, these roles can be unified in a single LLM to complete the procedure (namely \ourssft).
The results compared with methods in fine-tuning LLMs class demonstrate its superiority. 
Besides, the focus of \ours lies in exploiting inherent abilities of \llm, making it orthogonal to other solution categories and positioning it as a seamless integration into current research endeavors (ref. Section~\ref{sec:compatibility}).

To summarize, our contributions in this work are as follows:
\begin{itemize}
	\item To the best of our knowledge, we are the first to explore the inherent ability of \llm in the line of leveraging mistakes;
    \item We introduce the idea of human psyche and propose a novel paradigm for \llm, namely \ours, comprising three roles, i.e., \mi, \mii, and \miii, to achieve self-denial of \llm for further accurate generation;
	\item Extensive experiments demonstrate that the proposed method outperforms competitors in aforementioned three categories, and is effective in addressing the misleading problem.
\end{itemize}

\section{Related Work}
\label{sec:rel}

This section briefs relevant efforts from three perspectives.

\begin{figure*}
	\centering
        \vspace{-1ex}
	\includegraphics[width=\linewidth]{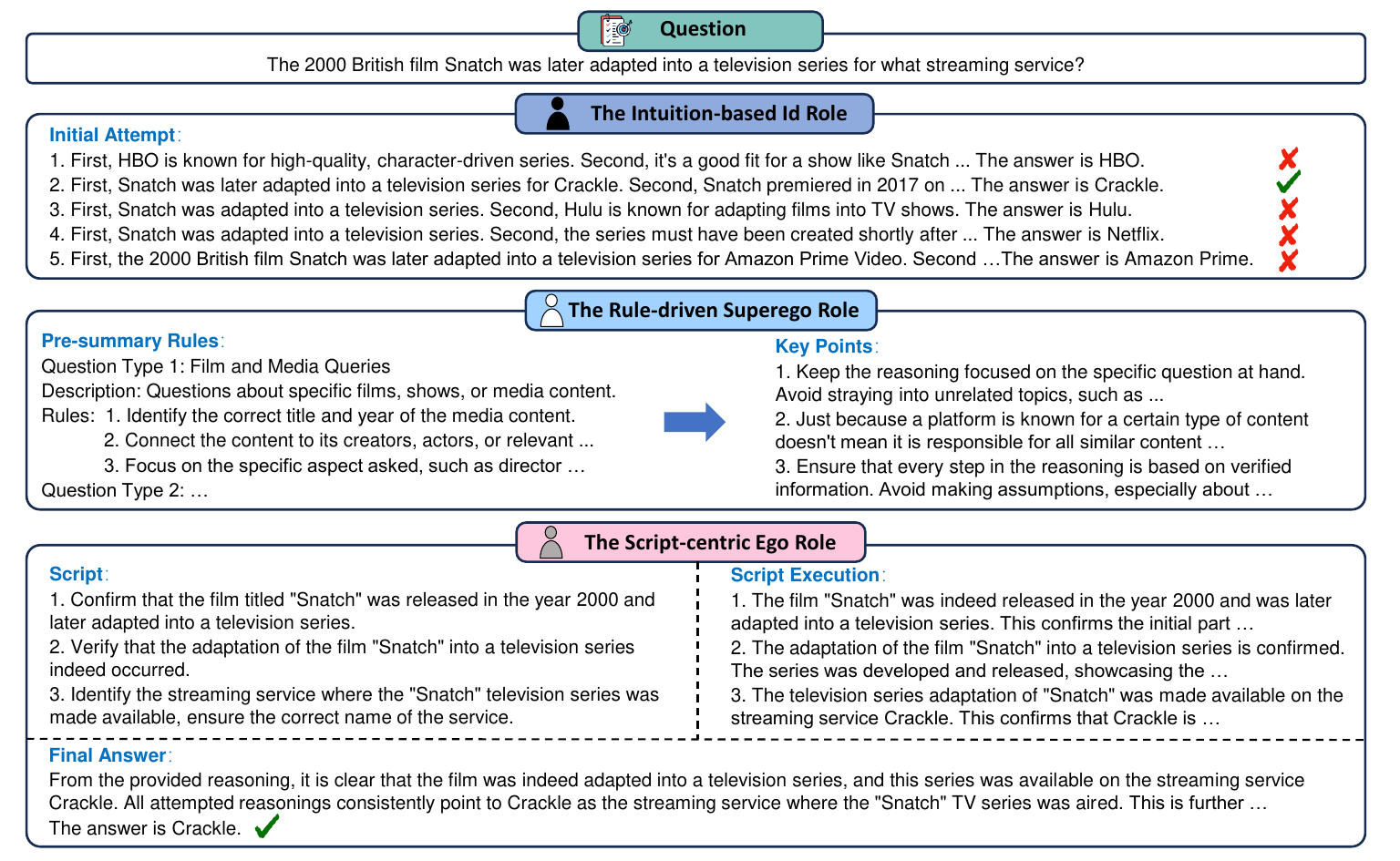}
	\vspace{-4ex}
	\caption{The framework of \ours. It is mainly constituted of the id role, the superego role, and the ego role. Here we expect that the discussion of the three roles will  enlighten LLMs to reason better.}
 \label{fig:framework}
 \vspace{-3ex}
\end{figure*}

\subsection{Fine-tuning \llm}
Methods in this approach treat \llm as a supervisor to obtain feedback for a provided mistake to fine-tune \llm.
\sta~\cite{DBLP:conf/nips/ZelikmanWMG22} emphasizes the rationales leading to the correct results. With the gold annotations, an iterative generation strategy is employed to obtain the ideal reasoning path for each question to fine-tune \llm.
\lema~\cite{DBLP:journals/corr/abs-2310-20689} focuses on inaccurate reasoning paths, and employs GPT-4 to generate details, reasons, and final answer for them. These samples then perform as annotations for fine-tuning \llm.
Similar to \lema, \gwfs~\cite{DBLP:journals/corr/abs-2310-10477} directly guides \llm to generate harmful responses and informs \llm to evaluate its output with specific critique. This mistake analysis data is then used for model fine-tuning. 
\srethink~\cite{DBLP:conf/acl/TongLWWTS24}pre-defines some error classes to provide \llm with typical correction examples to prepare fine-tuning samples.
\salam~\cite{DBLP:conf/emnlp/WangL23} designs an assistant LLM to interact with the main LLM and leverages the resulting conversations to fine-tune assistant LLM, thereby enhancing the supervisor's flexibility compared with aforementioned approaches.
The utilization of domain fine-tuning can enhance the performance of \llm; however, a new training phase encounters delayed response issue. 
Additionally, a substantial amount of computational powers are required to execute the learning process. 

\subsection{Leveraging Tools}
The main idea in this line is to leverage the feedback to verify \llm' outputs.
\tran~\cite{DBLP:conf/emnlp/YangLL23} accumulates rules from incorrect cases summarized by \llm to form a rule base. When encountering new queries, it extracts relevant rules as external clues for the reasoning process. 
\ve~\cite{DBLP:conf/acl/ZhaoLJQB23} transforms the attention to leverage external knowledge, e.g., Wikipedia and Google. The information performs as the facts to verify the results and supports the re-answer procedure.
To put the thoughts deeply, {ReAct}~\cite{DBLP:conf/iclr/YaoZYDSN023} proposes an interaction framework to enforce \llm to execute the thought-action-observation circle based on goolge engine. In each phrase, the model will make a feedback-dependent decision and prepare for the next state.
{Reflexion}~\cite{DBLP:conf/nips/ShinnCGNY23} tries reinforcement learning for reflection to further explore decision-making capability.
External assistance can help users to obtain immediate results; however, the performance is heavily reliant on the quality of knowledge source or effective tools. 
Once the assistance is inaccessible, the essential components of these methods are compromised.

\subsection{Multi-agent Debate}
Several agents are defined to invoke discussion within the same topic.
LM vs LM~\cite{DBLP:conf/emnlp/CohenHGG23} facilitates a multi-turn conversation between the claim-generating LM and an examining LM, which introduces questions to discover inconsistencies.
Multiagent Debate~\cite{DBLP:journals/corr/abs-2305-14325} makes multiple agents propose and debate their individual responses and reasoning processes over multiple rounds to reach a common final answer.
MAD~\cite{DBLP:journals/corr/abs-2305-19118} facilitates the expression of arguments among multiple agents in a "tit for tat" manner, while a judge oversees the debate process to obtain a final solution.
Self-Contrast~\cite{DBLP:conf/acl/ZhangSWPWZ024} employs diverse agents     to offer distinct perspectives on a query. Subsequently, a new agent compares and summarizes discrepancies to generate the final answer. 
The multi-agent debate framework effectively ensures timeliness and independence, but the omission of agents' self-denial results in perpetual resource competition.
\ours tries to tackle the identified challenges via the initial attempts, the regular guidance, and detailed executable steps from tailored roles.

\section{Proposed Method}
\label{sec:MRAF}
This section presents an overview of the proposed method and detailed design of each role.

\subsection{Framework}
As response to identified challenges, we follow the idea of human psyche and design three tailored roles. The overall framework of \ours is illustrated in Figure \ref{fig:framework}. 

For a given question, the \mi role relies on \llm' (e.g., GPT-4) inherent abilities to generate multiple reasoning paths (ref. Section~\ref{sec:mi}). 
Subsequently, these intuitive attempts serve as fundamental materials for the \mii to judge and criticize.
To enhance reliability of LLMs' self-supervision determination, we introduce general persuasive rules summarization based on the training samples from a methodological perspective.
Considering that even for human beings it is challenging to definitively assert the correctness of the original answer, these rules mainly help the agent in abstracting guiding key points based on the specific question and its associated attempts (ref. Section~\ref{sec:mii}). 
Third, the \miii role comprehensively synthesizes question, original results, and key points to construct a script.
The script provides step-by-step guidance for refining the potential mistakes in the reasoning process, thus facilitating the generation of the final answer (ref. Section~\ref{sec:miii}).\footnote{We provide all prompts of \ours on an GitHub repository: \url{https://github.com/liaozenghua/PSSD.git}.}
Last but not least, according to the theory, the functions of above three roles possess sequential relevance.
Consequently, we apply open-source LLMs (e.g., LLaMa and Mistral) to fine-tune these roles into an integration (ref. Section~\ref{sec:MRAF-SFT}). 

\subsection{The Intuition-based Id Role}
\label{sec:mi}
The \mi role performs as an initial reasoning attempt on a given question by leveraging the LLM’s inherent abilities, as defined by \citet{freud1923das}, to capture its intuitive responses as a foundation for subsequent analysis. 

Specifically, given a question $q'$, we provide the LLM with a id role prompt (i.e., $\mathcal{M}_{Id}$) with several examples as below:
$$\mathbf{E}=\{\mathbf{E}_i\}_{i=1}^{z}=\{(q^{t}_i,a^{t}_i)\}_{i=1}^{z}  ,$$
where $z$ denotes the number of selected examples, $q^{t}_i$ and $a^{t}_i$, used for few-shot learning, denote the $i$-th question along with its labelled answer from the training dataset.

We set the number of returned responses to $l$, and then $\mathcal{M}_{Id}$ generates the initial attempts, formally,
\begin{equation}
    \mathbf{A}=\{a_j\}_{j=1}^l=\mathcal{M}_{Id}(\mathbf{E},q').
\end{equation}

As shown in Figure \ref{fig:framework}, the \mi role generates multiple reasoning paths based on the LLM's basic capabilities, wherein the quality and accuracy of results vary randomly.
Through carefully comparing the similarities and differences among these paths, some delaying relevance might be obtained as the foundational reference for the next role.

\subsection{The Rule-driven Superego Role}
\label{sec:mii}

The \mii, representing internalized values and norms \cite{freud1923das}, is responsible for evaluating and regulating reasoning paths $\textbf{A}$ to ensure that specific key points are accurately acquired. 
To strengthen the reliability of this process, we abstract a set of highly relevant rules as a guideline to support generating key points.
These rules guide LLMs in identifying key points across various question types, thus providing a systematic framework that transcends individual responses. 
This framework helps LLMs overcome their limitations, especially when reasoning falls outside their factual knowledge base.

\subsubsection{Development of Rules}
A well-constructed set of rules is critical for effective key point generation. 
To develop these rules, we employ a contrastive approach using GPT-4. 
For each question $q^{t}_b$ in the training set, we instruct {GPT-4} to generate high-quality key points $K^h_b$, which are then manually reviewed for accuracy. 
In parallel, we instruct {LLaMA-2-7B-Chat} to generate suboptimal key points $K^s_b$. 
The comparison between these two sets of outputs provides a gradient that allows {GPT-4} to learn what constitutes a high-quality key point.
This process is divided into two stages: pattern extraction and rule summary.

\paragraph{Pattern Extraction}
For each question $q^{t}_b$, we instruct {GPT-4} to contrast high-quality key points $K^h_b$ with suboptimal ones $K^s_b$ to identify meaningful patterns $P_b$. 
These patterns explain the distinctions between the two key points, mathematically,
\begin{equation}
\begin{aligned}
P_b &= {\arg \max_b} \ \ \mathrm{P}\left(\mathcal{T} \mid q^{t}_b, K^h_b,K^s_b\right)\\
& =\left\{t_{1}, t_{2}, \cdots, t_{m}\right\},
\end{aligned}
\end{equation}
where $\mathcal{T}$ represents all possible patterns, and the top $m$ patterns are selected. 
We extract patterns for each question in the training set, and this process results in a pattern set $\mathbf{P}=\{P_1,P_2,\cdots,P_{|P|}\}$.

\paragraph{Rule Summary}
Based on extracted pattern set $\textbf{P}$, we instruct {GPT-4} to categorize all questions $\mathbf{Q}^{t}=\{q^{t}_1,q^{t}_2,\cdots,q^{t}_{|\mathbf{Q}^{t}|}\}$ into several types and generate corresponding rules for each type.
These rules describe the key insights derived from patterns across similar questions, which can guide LLMs from a methodological perspective to better generate key points, formally,
\begin{equation}
\begin{aligned}
U &= {\arg \max_n} \ \mathrm{P}\left(\mathcal{U} \mid \mathbf{Q}^{t}, \mathbf{P}\right)\\
& =\left\{u_{1}, u_{2}, \cdots, u_{n}\right\},
\end{aligned}
\end{equation}
where $\mathcal{U}$ represents the rule set for all possible question types. 
The rules summarized by this method are used to guide LLMs in generating key points.\footnote{Regarding hyper-parameter $m$ and $n$, we first establish parameter ranges following related studies \citep{DBLP:conf/acl/ZhouZY0YCN22}. 
We then refine the values slightly and select the optimal parameters ($m=3$ and $n=10$) according to experimental results.}

\subsubsection{Key Point Generation}
Using a predefined superego role prompt, i.e., $\mathcal{M}_{Superego}$, we require the LLM to analyze reasoning attempts $A$ and generate key points $K$ for a given question $q'$ under the guidance of rule $U$ as below: 
\begin{equation}
    K=\mathcal{M}_{Superego}(U,q',\mathbf{A}).
\end{equation}

As illustrated in Figure \ref{fig:framework}, the \mii role evaluates initial reasoning attempts, identifying incorrect logic (e.g., key point 2 in attempt 1) and guiding corrections in subsequent steps.

\subsection{The Script-centric Ego Role}
\label{sec:miii}
The \miii role tries to mediate conflicts between the \mi role and \mii role by striking a balance between the original attempts and the obtained key points.
This process embodies the methodological guidance into a specific executable script that illustrates detailed steps for refining previous results, which briefly involves three main steps. 

\subsubsection{Script Generation}
In detail, we provide a LLM with the predefined ego role prompt (i.e., $\mathcal{M}_{Ego}$) to analyze  the initial attempt $\textbf{A}$ and key points $K$. 
The step-wise script $S$ is generated as below:
\begin{equation}
    S=\mathcal{M}_{Ego}(q',\textbf{A},K).
\end{equation}

The script synthesizes LLMs' reasoning capabilities with the summarized rules, distinguishing from key points in terms of levels. 
As depicted in Figure~\ref{fig:framework}, key points emphasize what to do for enhancing the precise of the answer, such as ``Ensure that every step in the reasoning is based on verified information'',
The script focuses on how to achieve these targets like ``Verify that the adaptation of the film `Snatch' into a television series indeed occurred''.

\subsubsection{Script Execution}
In this step, the \miii role executes the script step-by-step, ensuring that LLMs follow the reasoning path outlined in the generated script.
In order to keep the execution process smoothly, we integrate the script with the key points as input to guide $\mathcal{M}_{Ego}$, formally, 
\begin{equation}
    S'=\mathcal{M}_{Ego}(q',K,S),
\end{equation}
where $S'$ denotes the answered script as illustrated in Figure~\ref{fig:framework}.

After the execution, the script further confirms details relevant to the question and provides supplementary descriptions.
These references contribute to the accuracy of the final decision.

\subsubsection{Answering}
Finally, the \miii role formulates the final answer based on all relevant process information generated in the previous steps. 
Specifically, we instruct the LLM to synthesize the question $q'$, key points $K$, script $S$, and script execution $S'$ to generate the final result $R$.
This process is formalized as: 
\begin{equation}
    R=\mathcal{M}_{Ego}(q',K,S,S').
\end{equation}

\subsection{The Merge of Three Roles}
\label{sec:MRAF-SFT}
To mitigate resource competition in multi-agent debate, we further explore effective management strategy for frequency of invoking LLMs.
In accordance with the internal unity of three roles, they are merged into a whole through the utilization of open-source LLMs.
As fine-tuning process performs as an adhesive, combining distinct roles, the modified method is called \ourssft to differentiate it from \ours.
Subsequently, we discuss the construction of the training dataset, along with the training and reasoning processes involved in \ourssft in greater detail.

\subsubsection{Construction of Training Dataset}
First of all, the fine-tuning process is essential a supervised learning procedure, thereby necessitating annotated samples as training labels.
To construct required dataset, 
we use publicly available reasoning datasets as the data source.\footnote{For textual reasoning, we select \advhot and \wiki as data source, while for mathematical reasoning, we choose \gsm and \mat as data source.} 
For each question $q^{t}_b$ in training set, we apply GPT-4 as the fundamental model of \ours framework to generate reasoning records, which include the initial attempts, key points, script, script execution, and final answer. 

To ensure accuracy, these reasoning records undergo a rigorous annotation process:
\begin{itemize}
    \item Initial Annotation. Three graduate students initially annotate these reasoning records and carefully review and improve them based on the groundtruth. 
    \item Review and Consensus. Each record is thoroughly reviewed by an additional annotator. 
    Any discrepancies are thoroughly discussed until a consensus is reached on the final annotation.
    \item Structured Combination. Each annotated reasoning record is integrated into a predefined training sample template.  
\end{itemize}

Finally, 
This process can be formally defined as follows:
$$d_b=Template[q^t_b\oplus \mathbf{A}_b\oplus K_b\oplus S_b\oplus S'_b\oplus R_b],$$
where $q^{t}_b$, $\mathbf{A}_b$, $K_b$, $S_b$, $S'_b$ and $R_b$ denote the given question, corresponding initial attempts, key points, script, script execution and final answer respectively. $\oplus$ denotes the operation of concatenation.

\subsubsection{Fine-Tuning LLMs}
To further conserve computational resources, we employ the LoRA~\cite{DBLP:conf/iclr/HuSWALWWC22} method for fine-tuning the LLMs. 
On one hand, considering the \mi role purely relies on the capability of the fundamental model, its enhancement emphasizes the improvement of LLMs.
Therefore, we fine-tune the first LoRA model using question-reasoning path pairs as training data. 
This process can be formulated as:
\begin{equation}
W_1: \mathcal{L}_1=-\frac1L\sum_{i=1}^{L}\log P(x_i|q^t_b,a_b^{<i}),
\end{equation}
where $W_1$ represents weights of the first LoRA model, $\mathcal{L}_1$ represents its loss function, $L$ is the token length of the sequence $a_b$, $x_i$ is the currently predicted response token, $q^t_b$ and $a_b^{<i}$ are the question and the response tokens before $x_i$.

On the other hand, acquired structured  $\textbf{D}=\{d_1,d_2,\ldots,d_{|D|}\}$ aims to provide supervised signals for training other more complicated roles (i.e., \mii and \miii).
Thus, it is employed to fine-tune another LoRA model focusing the generation of elements in $d$, formally,
\begin{equation}
W_2: \mathcal{L}_2=-\frac1L\sum_{i=1}^{L}\log P(x_i|q^t_b,\mathbf{A}_b,d_b^{<i}),
\end{equation}
where $\mathbf{A}_b$ and $d_b^{<i}$ are the initial attempt with 5 reasoning paths and the response tokens before $x_i$.

This fine-tuning strategy strengthens LLMs not only in generating diverse reasoning paths during initial attempts but also in developing multi-role adaptive reasoning capabilities.

The parameters of these two LoRA models are ultimately merged into the original fundamental model, indicating that only one model supports the operation of \ourssft, mathematically,
\begin{equation}
W = W_0 + W_1 + W_2
\end{equation}
where $W$ and $W_0$ represent the weights of the merged model $\mathcal{M}_{LoRA}$ and the original weights of the LLM.

Specifically, the complete reasoning process in \ourssft involves two main steps.
For a given input question $q'$, \ourssft first generates $l$ initial reasoning paths $\textbf{A}$.
Then, using both $q'$ and $\textbf{A}$ as inputs, it sequentially generates the key points $K$, script $S$, script execution $S'$, and finally produces the final answer $R$, as below:
\begin{equation}
\mathcal{M}_{LoRA}: q' \rightarrow  \textbf{A} \rightarrow  K,S,S',R.
\end{equation}
where $\mathcal{M}_{LoRA}$ denotes the merged LoRA model with weight $W$.

\section{Experiments}
\label{sec:exp}

This section provides a detailed presentation of the experiments, along with an in-depth analysis of the results.

\begin{table*}[!t]
\caption{Overall results (\%) of \ours on GPT-4. $\triangle$ denotes the improvement from \cotsc baseline. The best results in each dataset are highlighted in bold. AvgT denotes the average time (in seconds) required for each individual reasoning process. - denotes the corresponding information does not exist. 
 Considering LLMs make only a single attempt in both the \stand and \cotori settings, the PM value is not applicable in these cases. 
 Since external knowledge bases utilized by \react and \ve cannot support math computation, they are not conducted in mathematical reasoning datasets. Details in Section~\ref{sec:eva-MRAF}.}\vspace{-2ex}
	\centering \small
	\setlength{\tabcolsep}{5mm}{
		\begin{tabular}{c|l|c|c|c|c|c|c}
            \toprule
			\textbf{Dataset} & \textbf{Method} &  \textbf{EM} & $\triangle$ \textbf{EM} & \textbf{PM} & $\triangle$ \textbf{PM} & \textbf{RM}   &  $\textbf{AvgT}$ \\
			\midrule
			\multirow{7}{*}{\advhot} 
			& \stand &  36.36 & - & - & - & -  & 1.81 \\
			& \cotori &  42.86 & - & - & - & -  & 2.89\\
			& \cotsc &  42.53 & - & 72.73 & - & 0.00  & 4.31\\
			& \react  & 44.81 & +2.28 & 59.46 & -13.27 & 10.97  & 14.08 \\
			& \ve &  44.16 & +1.63 & 60.81 & -11.92 & 7.74 & 23.78\\
   & \scontrast &  45.13 & +2.60 & 68.53 & -4.20 & 9.84 & 34.64\\
   & \ours &   \textbf{47.08} & +\textbf{4.55} & \textbf{74.32} & +\textbf{1.59} & \textbf{11.08} & 19.80\\
			\midrule  
			\multirow{7}{*}{\wiki} 
			& \stand & 34.82 & - & - & - & - & 1.80 \\
			& \cotori &37.20 & - & - & - & - & 2.92 \\
			& \cotsc &  40.18 & - & 68.37 & - & 0.00 & 4.11 \\
			& \react &  \textbf{45.24} &\textbf{+5.06} & 65.84 &-2.53 & \textbf{14.32} & 16.09\\
			& \ve &  42.26 & +2.08 & 62.16 & -6.21 & 11.36 & 20.73\\
   & \scontrast &  41.37 & +1.07 & 70.67 & +2.30 & 9.64 & 29.45 \\
			& \ours &  41.96  & +1.78 & \textbf{72.43} & +\textbf{4.06} & 9.48 & 19.14\\	
   		\midrule	
			\multirow{5}{*}{\gsm} 
			& \stand & 88.00 & - & - & - & - & 2.03\\
			& \cotori &93.60 & - & - & - & - & 4.87\\
			& \cotsc &  93.20 & - & 76.27 & - & 0.00 & 7.68 \\
   & \scontrast &  95.40 & +2.20 & \textbf{86.71} & +\textbf{10.44} & 10.00 & 49.58 \\
   & \ours  &   \textbf{96.80} & +\textbf{3.60} & 84.75 & +8.48 & \textbf{22.22} & 34.14 \\
      		\midrule
			\multirow{5}{*}{\mat} 
			& \stand & 65.20 & - & - & - & - & 2.38\\
			& \cotori &73.40 & - & - & - & - & 5.34\\
			& \cotsc &  72.80 & - & 70.53 & -  & 0.00 & 7.83 \\
   & \scontrast &  76.00 & +3.20 & 78.36 & +7.83 & 8.92 & 53.92\\
   & \ours  &   \textbf{77.20} & +\textbf{4.40} & \textbf{80.18} & +\textbf{9.65} & \textbf{13.17} & 36.41 \\
			\bottomrule
	\end{tabular}
 }

	\label{tab:mraf}
    \vspace{-3ex}
\end{table*}

\subsection{Experiment Setup}
\subsubsection{Datasets}
We conduct experiments on both textual and mathematical reasoning tasks using four datasets: \advhot, \wiki, \gsm, and \mat.

For textual reasoning, we utilize the following datasets: (1) \advhot \footnote{\url{https://github.com/xiye17/TextualExplInContext}.}
~\cite{DBLP:conf/nips/YeD22} is a challenging subset derived from the multi-hop question answering dataset \textsc{HotpotQA}~\cite{DBLP:conf/emnlp/Yang0ZBCSM18}, where the correct and incorrect predictions are balanced;
(2) \wiki
 \footnote{\url{https://github.com/Alab-NII/2wikimultihop}.}
 ~\cite{DBLP:conf/coling/HoNSA20} is a multi-hop question answering dataset that leverages the structured format of Wikidata and applies logical rules.

For mathematical reasoning, we use:
(1) \gsm \footnote{\url{https://github.com/openai/grade-school-math}.}
 ~\cite{DBLP:journals/corr/abs-2110-14168}, a benchmark dataset of grade-school-level math word problems, designed to evaluate mathematical reasoning. Each question is accompanied by a detailed step-by-step solution.
(2) \mat \footnote{\url{https://github.com/hendrycks/math}.}
 ~\cite{DBLP:conf/nips/HendrycksBKABTS21} is a collection of challenging high school-level math problems, aimed at evaluating advanced mathematical reasoning and problem-solving skills.

\subsubsection{Competitors}
To put our results in perspective, we apply two classes of competitive baselines for demonstration.

In the first class, we select representative methods from leveraging-tools and multi-agent debate categories to evaluate \ours, as these methods commonly employ closed-source LLMs (e.g., GPT-4) as the fundamental model:
(1) Standard Prediction (\stand) directly predicts the answer for input using manually provided examples; 
(2) Original \cott (\cotori)~\cite{DBLP:conf/nips/Wei0SBIXCLZ22} generates a reasoning path before predicting the final answer;
(3) \cott with Self-Consistency (\cotsc)~\cite{DBLP:conf/iclr/0002WSLCNCZ23}samples five reasoning paths and selects the final answer based on consistency values;
(4) \react~\cite{DBLP:conf/iclr/YaoZYDSN023} enhances CoT reasoning by integrating Wikipedia API to improve factual accuracy;
(5) \ve~\cite{DBLP:conf/acl/ZhaoLJQB23} is a post-editing framework that leverages external knowledge to increase the factual accuracy of predictions;
(6) Self-Contrast~\cite{DBLP:conf/acl/ZhangSWPWZ024} is a self-contrast framework that compares multiple solution perspectives by multiple agents to re-examine and eliminate mistakes.

In the second class, we compare \ourssft with the following state-of-the-art fine-tuning methods based on the same open-source LLMs (e.g., LLaMA):
(1) CoT Fine-tuning fine-tunes the target model using the CoT reasoning paths from the target dataset;
(2) Mistake Tuning~\cite{DBLP:conf/acl/TongLWWTS24} fine-tunes the model using both correct and incorrect reasoning paths to improve error correction;
(3) AugGPT~\cite{DBLP:journals/corr/abs-2302-13007} generates training data using the specific sample as a seed, and then fine-tunes the target model on the generated data; 
(4) LEC~\cite{DBLP:journals/corr/abs-2407-00497} utilizes error-prone samples from the target model as seeds to generate training data, followed by fine-tuning on the data.

All experiments adhered to the default hyper-parameters reported in their papers. 
Fine-tuning experiments are conducted using the LoRA method to optimize computational efficiency.

\subsubsection{Metrics}
We evaluate the performance of \ours via three metrics: 
(1) Exact Match (EM) measures the percentage of predictions that exactly match the ground truth, which evaluates the answering ability of models; 
(2) Potential Match (PM) measures the conditional probability of accurately selecting the correct answer from original attempts wherein the gold answer already exists, which evaluates the ability to distinguish between correct and incorrect results;
(3) Rectified Match (RM) measures the conditional probability of original incorrect predictions that is finally corrected, which evaluates the capability to correct.

\subsubsection{Implementations}
To evaluate the effectiveness of \ours and \ourssft, we select both leading closed-source and powerful open-source LLMs as baselines.
For closed-source LLMs, we use GPT-4-turbo~\footnote{We use GPT-4-turbo-2024-04-09 as default GPT-4-turbo version}~\cite{DBLP:journals/corr/abs-2303-08774},
accessed via its APIs.
For open-source LLMs, select the powerful open-source LLMs Mistral-7B-Instruct-v0.2~\footnote{\url{https://huggingface.co/mistralai/Mistral-7B-Instruct-v0.2}} and LLaMA-3-8B-Instruct~\footnote{\url{https://huggingface.co/meta-llama/Meta-Llama-3-8B-Instruct}} as baseline models.
In order to ensure fairness in our experiments, the same foundation is utilized to support competitors for the comparison.  
The experiments are performed with 2 * RTX 3090.
During the training phase of \ourssft, we utilize the AdamW optimizer \citep{DBLP:conf/iclr/LoshchilovH19} with $\beta_1=0.9$ and $\beta_2=0.999$. 
The learning rate is set to 1e-6, with a 0.1 ratio of warm-up steps and linear decay. 
We configure the maximum input length to 4,096 tokens and establish a training batch size of 4. 
The entire training process is completed within 4.5 hours, and we employ the final checkpoint for subsequent evaluations.

\vspace{-1ex}

\subsection{\ours Results}
\label{sec:eva-MRAF}
As shown in Table~\ref{tab:mraf}, \ours consistently achieves state-of-the-art results across nearly all tasks in all metrics, demonstrating the superiority and generalizability of our design.
The methods, such as \stand, \cotori, and \cotsc, only employ LLMs without any additional architectural modifications.
Thus, their average reasoning time exhibits a low value and should be considered solely for reference purposes.
Interestingly, the introduction of \cotsc does not significantly improve the EM value over \cotori; in fact, it reduces the EM value on three datasets, except for \wiki. 
This outcome aligns with human intuition, as \cotsc merely repeats the original inference multiple times without considering underlying mistakes. 
These facts demonstrate the advantage of distinct psychological roles of \ours on enhancing accuracy of the reasoning, wherein it achieves 4.00\% increase over \cotori in EM on average (for simplicity we will omit the “average” in the follows if it refers to the mean value of all accessible datasets).

In both \react and \ve, \cotsc is applied to generate consistency values for each question, and then external knowledge base rectifies the filtered generations.
The direct application of Wikipedia can account for the 4.17\% and 0.3\% higher EM performance of \react and \ve compared to \ours in \wiki.
The comparison of ranking positions in \advhot and \wiki further illustrates that the performance of methods in leveraging tools heavily relies on the quality of external facts.
Once the supporting references cannot meet the requirement, their performance experiences a cliff decline, as results in \advhot.
Even so, \ours outperforms \react by 10.75\% and \ve by 11.89\% in PM, indicating its effectiveness in stimulating LLMs for mistake awareness.
And the higher values in RM (except for \wiki) of \ours demonstrate the correcting ability of our designs. 

\ours beats \scontrast by 1.32\% in EM, which is in accordance with our anticipation, since this typical multi-agent debate merely assigns several roles to discuss and generate the answer.
In comparison with \ours, without consideration of the self-denial, it falls short on distinguishing correct generations from incorrect ones (a 1.85\% drop in PM) and rectifying original total wrong attempts (a 4.39\% drop in RM).
When it comes to the mentioned resource competition, \ours gets the reduction reasoning time by 14.7s below \scontrast, saving the time cost. 
The number of agent (3) and times of evoking LLMs (5) are also superior to those of \scontrast (4, 7.8), as listed in Table~\ref{tab: resource}.
These findings strongly supports the low-resource requirements of our three psychological tailored roles.

As aforementioned, the detailed analysis of the comparison with approaches in fine-tuning LLMs is provided in Section~\ref{sec:eva-MRAF-SFT}.

\begin{table}[!t]
\caption{
The results in EM (\%) of \ourssft. \textbf{\textsc{Adv.}} means \advhot dataset and \textbf{\textsc{Wiki.}} means \wiki dataset. The best results in each dataset are highlighted in bold, while the second positions are underlined. Details in Section~\ref{sec:eva-MRAF-SFT}.}\vspace{-2ex}
\centering
\small
\tabcolsep=0.015
\linewidth
\begin{tabular}{L{0.23\linewidth}C{0.16\linewidth}C{0.10\linewidth}C{0.10\linewidth}C{0.13\linewidth}C{0.11\linewidth}}
\hline
\textbf{Method}            & \textbf{Model}      & \textbf{\textsc{Adv.}}                   & \textbf{\textsc{Wiki.}}                    & \textbf{\gsm}               & \textbf{\mat}                    \\ \hline
CoT Fine-Tuning           & \multirowcell{5}{Mistral-7B-\\Instruct}  &   30.84   &     30.06  &    58.00   &   41.80  \\
AugGPT & ~  &    \textbf{32.47}   &   29.46   &   54.40  &    \textbf{43.40}      \\
LEC    & ~ & 29.87 & 28.27 &54.00& 41.20\\ 
Mistake Tuning  & ~  &   30.19  &    \textbf{30.95}    & \underline{58.60}  &    \underline{43.20}    \\
\ourssft & ~  &      \underline{32.14}    & \underline{30.65}   &   \textbf{59.20}  &    \textbf{43.40}     \\ \hline
CoT Fine-Tuning  & \multirowcell{5}{LLaMA-3-\\8B-Instruct} &    31.50        &    \textbf{32.44}   &   79.50   &   42.20    \\
AugGPT & ~ &   \underline{32.14}      &  30.65   &     77.80 &   39.20     \\
LEC    & ~ & 30.19 & 28.86  &80.20 & 38.40 \\ 
Mistake Tuning  & ~ &    30.52   &  31.25     &  \underline{80.60}   &   \underline{43.80}   \\
\ourssft & ~  &    \textbf{32.47}    &    \underline{32.14}    &   \textbf{80.80}   &     \textbf{44.60}    \\ \hline
\end{tabular}

\label{tab:mraf-sft}
\vspace{-2ex}
\end{table}

\vspace{-1ex}
\subsection{\ourssft Results}
\label{sec:eva-MRAF-SFT}
As shown in Table~\ref{tab:mraf-sft}, \ourssft achieved competitive performance on all benchmarks, particularly excelling in mathematical reasoning, where it delivered state-of-the-art results. 

In comparison, the results of AugGPT and LEC indicate that additional LLM-generated training data does not always yield positive outcomes.
For instance, on \wiki and \gsm, this approach even underperformed compared to the dataset's built-in CoT training set. 
The result might be attributed to biases introduced by LLM-generated data, which can impair the fine-tuned LLMs' specific reasoning capabilities.
Mistake Tuning
shows promise by enhancing LLMs' ability to distinguish between correct and incorrect responses. 
However, the contrastive-style fails to approach LLMs' self-denial mechanism.
Therefore, the model merely learns superficial features of mistakes.

In contrast, \ourssft contains three psychological roles with their tailored fine-tuning methods (i.e., two stages), aiming to stimulate LLMs' self-denial.
The merged LoRA model proves its significance through the superior results. 
The outperformance, meanwhile, indicates that the integration of roles from \ours into an entity (smaller open-source LLMs) through fine-tuning is both a viable and effective strategy.
Furthermore, the resource-friendliness feature of our design is also validated according to the numbers of agents, debate rounds, and times of invoking LLMs (ref. Table~\ref{tab: resource}).

\subsection{Ablation Study}
\label{sec:ablation}

\begin{table}[!t]
\caption{Ablation studies on \ours in EM (\%) employing GPT-4.
\textbf{\textsc{Adv.}} means \advhot dataset. Id represents the \mi role, selecting the final answer through self-consistency. Superego-R represents the \mii role without using summarized rules to generate key points. Superego represents the full superego role. Ego represents the script-based reasoning of the \miii role. Details in Section~\ref{sec:ablation}}\vspace{-2ex}
\centering
\small
\tabcolsep=0.028\linewidth
\begin{tabular}{cccc|cc}
\hline
\textbf{Id} &  \textbf{Superego-R} &  \textbf{Superego} & \textbf{Ego} & \textbf{\textsc{Adv.}} & \textbf{\gsm} \\ \hline
\checkmark  &  \checkmark    &   \checkmark   &  \checkmark &   \textbf{47.08}  &  \textbf{96.80} \\ 
\hline
\checkmark   & \checkmark     & \checkmark     &   $\times$   & 45.45 & 95.60  \\
\checkmark    &   \checkmark   &   $\times$   &  $\times$   & 44.81  & 95.00  \\
 \checkmark      & $\times$ &   $\times$   &  $\times$    & 42.53 & 93.20  \\
 \hline
\end{tabular}

\label{tab: ablation}
\vspace{-3ex}
\end{table}

To analyze effects of different roles in \ours, we perform the ablation study and results is presented in Table~\ref{tab: ablation}. 
Due to the sequential feature of \ours, in each setting, we instruct LLMs to derive a final answer based on the information generated by remaining roles.

In comparison with the complete model, the removal of specific roles results in performance drop.
Notably, the whole \mii role provides the increase of 1.92\% in \advhot and 2.40\% in \gsm, which supports the significance of first identified challenge.
Once the model is able to confidentially and convincingly supervise its generation, it starts to engage in self-denial.
The fact that the introduction of summarized rules improve performance by 0.64\% in \advhot and 0.60\% in \gsm indicates a strongly positive correlation between persuasiveness of references and the accuracy of results.
Moreover, the equipment of the \miii role contributes to a 1.63\% improvement in \advhot and a 1.20\% improvement in \gsm, illustrating the value of handling the second challenge identified.
After discerning potential mistakes, the machine should further know targeted steps for rectifying initial attempts, which is effectively tackled by our script design. 

Overall, the study demonstrates the indispensable nature of roles in \ours, with each fulfilling a distinct and crucial function based on Freud’s psychological theory. 
The contribution of each role during the self-denial process differentiates, and the combine efforts ensure that \ours effectively enhances LLMs' reasoning capabilities.

\subsection{Compatibility Study}
\label{sec:compatibility}

\begin{table} [!t]
\caption{Compatibility Study of \ours in EM (\%). The best results in each dataset are highlighted in bold.$\uparrow$ denotes the improvement from \ours. Details in Section~\ref{sec:compatibility}.}\vspace{-2ex}
	\centering \footnotesize
	\setlength{\tabcolsep}{4mm}{
		\begin{tabular}{l|c|c}
			\toprule
			\textbf{Method} & \textbf{\advhot} & \textbf{\wiki} \\
			\midrule		  
            \react & 44.81& 45.24 \\
            \ve & 44.16& 42.26 \\
            \ours & 47.08& 41.96 \\
			\midrule
            \ours + \react & \textbf{48.05} ($\uparrow$ 0.97) & \textbf{45.54} ($\uparrow$ 3.58) \\
			\ours + \ve  & \textbf{48.05} ($\uparrow$ 0.97) & 42.86 ($\uparrow$ 0.90) \\			
			\bottomrule
	\end{tabular}}
	
    \vspace{-4ex}
    \label{tab:compatibility}
\end{table}

To demonstrate that \ours is seamlessly orthogonal to other category, we evaluate it by incorporating leveraging-tools methods, such as \ve and \react. 
During the final answer generation step in \ours, we set the number of returned candidate texts to $5$ and apply above methods to assist in answering questions when its consistency values fall below $3$.
The result is provided in Table~\ref{tab:compatibility}.

The results demonstrate that when combined with \ve or \react, \ours can significantly enhance the reasoning capabilities of LLMs.
This illustrates strong compatibility or \ours with external retrieval systems, thereby highlighting its adaptability and effectiveness in leveraging additional knowledge.
The fact suggests a future direction for further optimizing the proposed methodology.
Besides, the observed increase (i.e., 3.58\%) in \wiki combined with ReAct and the highest result substantiate our hypothesis that the leverage of Wikipedia determines the best result of ReAct in \wiki (ref. Section~\ref{sec:eva-MRAF}).

\begin{figure} 	
	\centering
        \vspace{-2.5ex}
	\subfigure[Results in \cotsc.]{
		\includegraphics[width=0.46\linewidth]{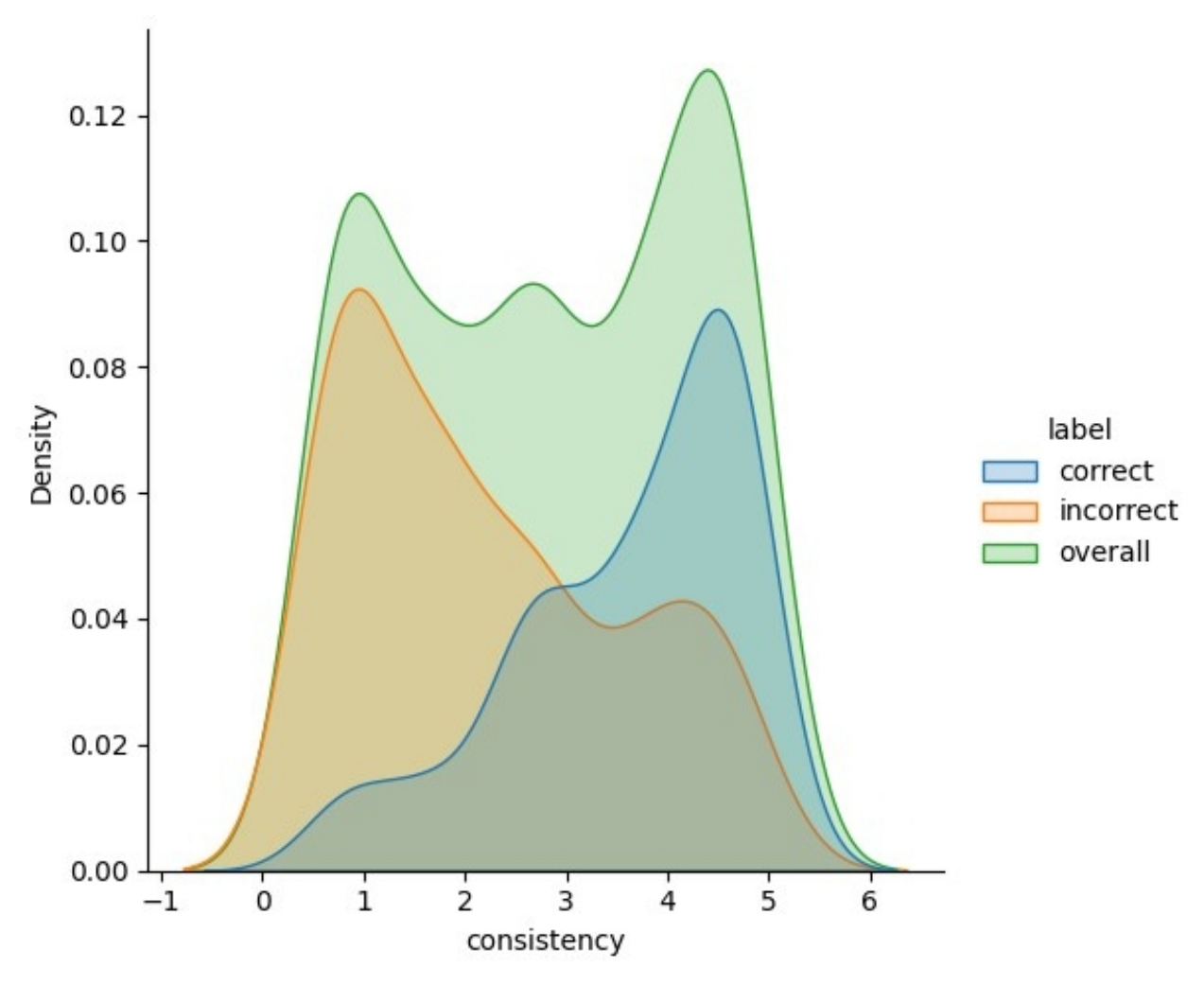}
	}
	\subfigure[Results in \react.]{
		\includegraphics[width=0.46\linewidth]{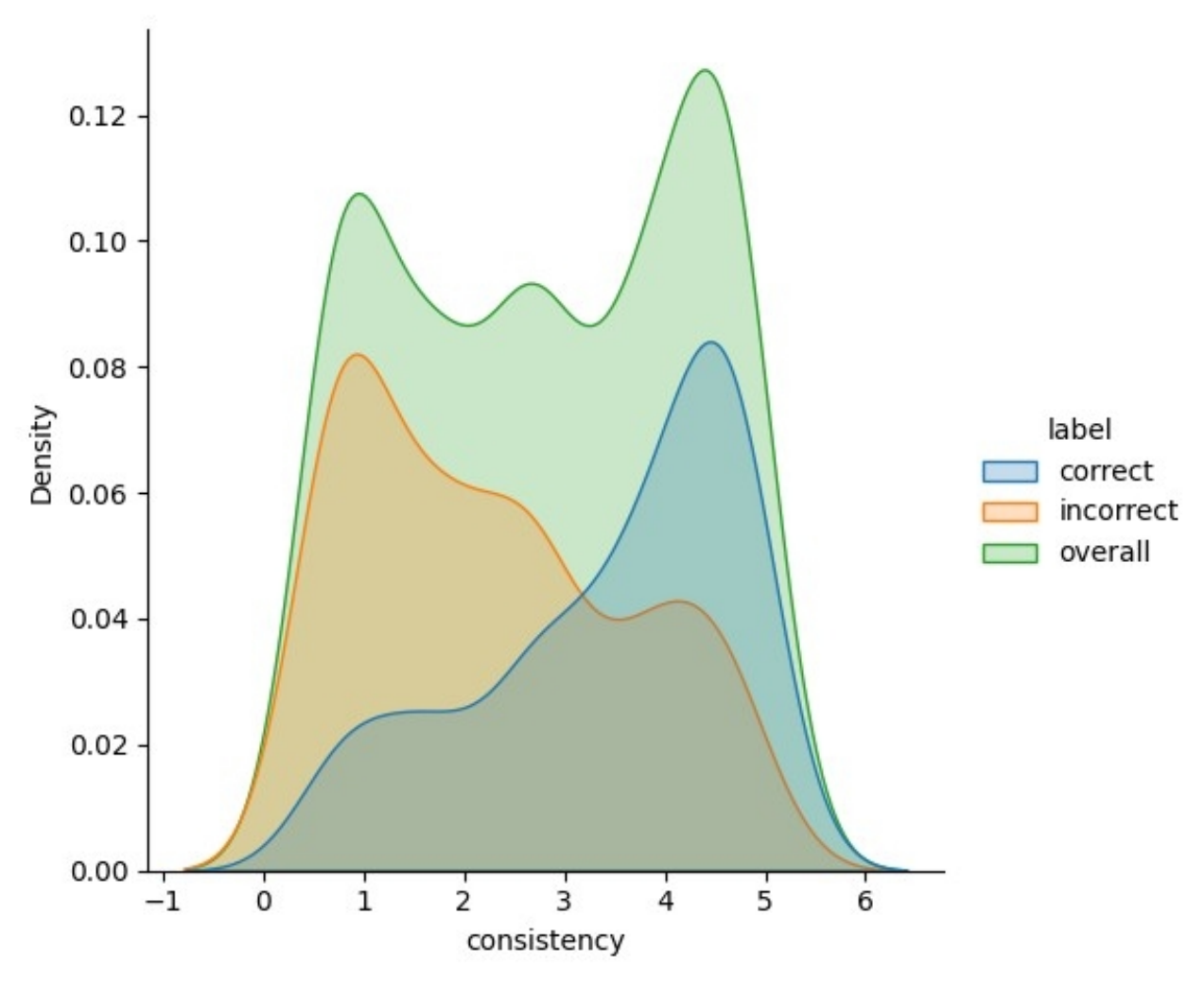}
	}
        \vspace{-2.5ex}
        \subfigure[Results in \ve.]{
		\includegraphics[width=0.46\linewidth]{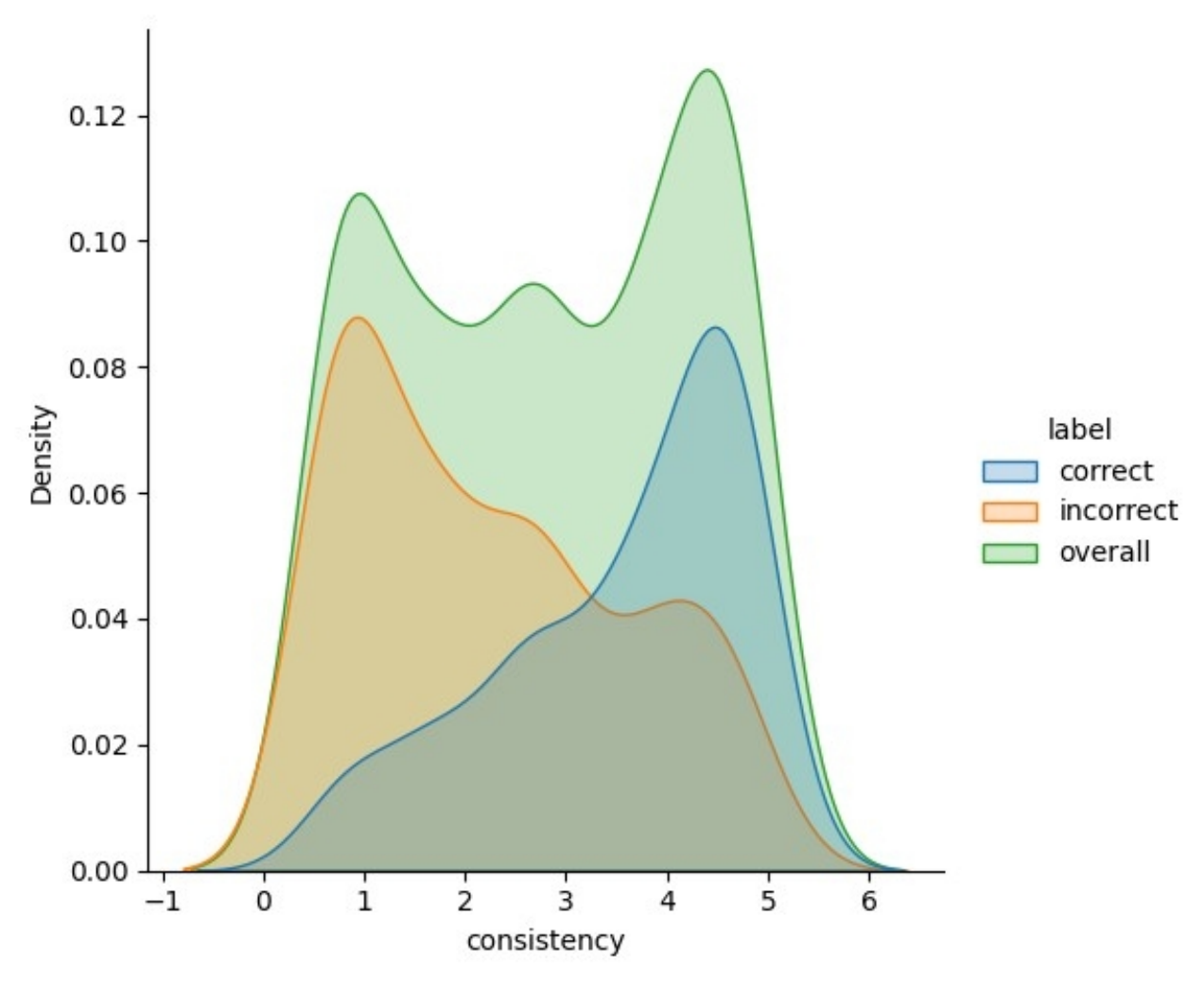}
	}
 	\subfigure[Results in \ours.]{
		\includegraphics[width=0.46\linewidth]{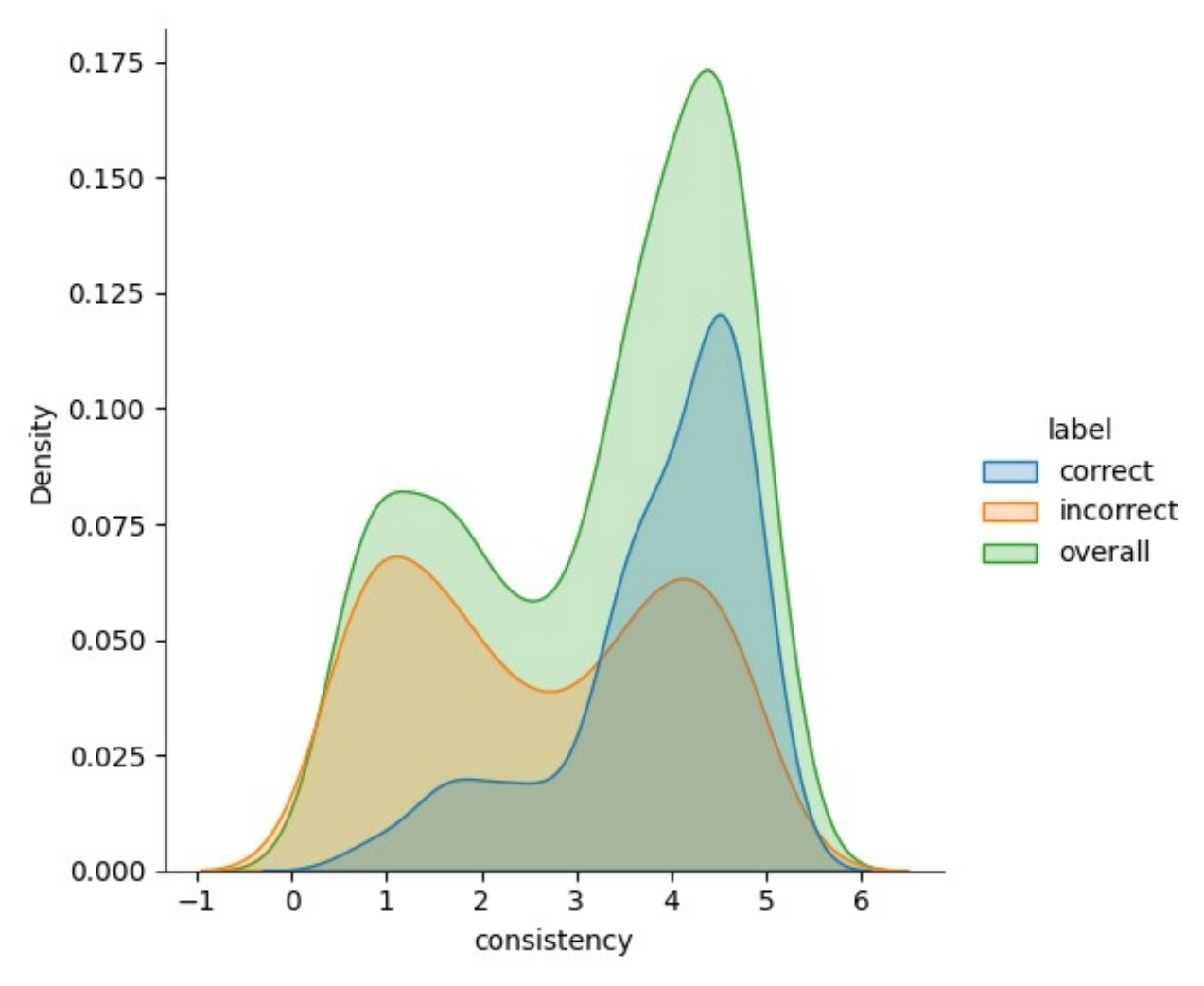}
	}
        \vspace{-2.5ex}
	\subfigure[Results in     \ours+\react.]{
		\includegraphics[width=0.46\linewidth]{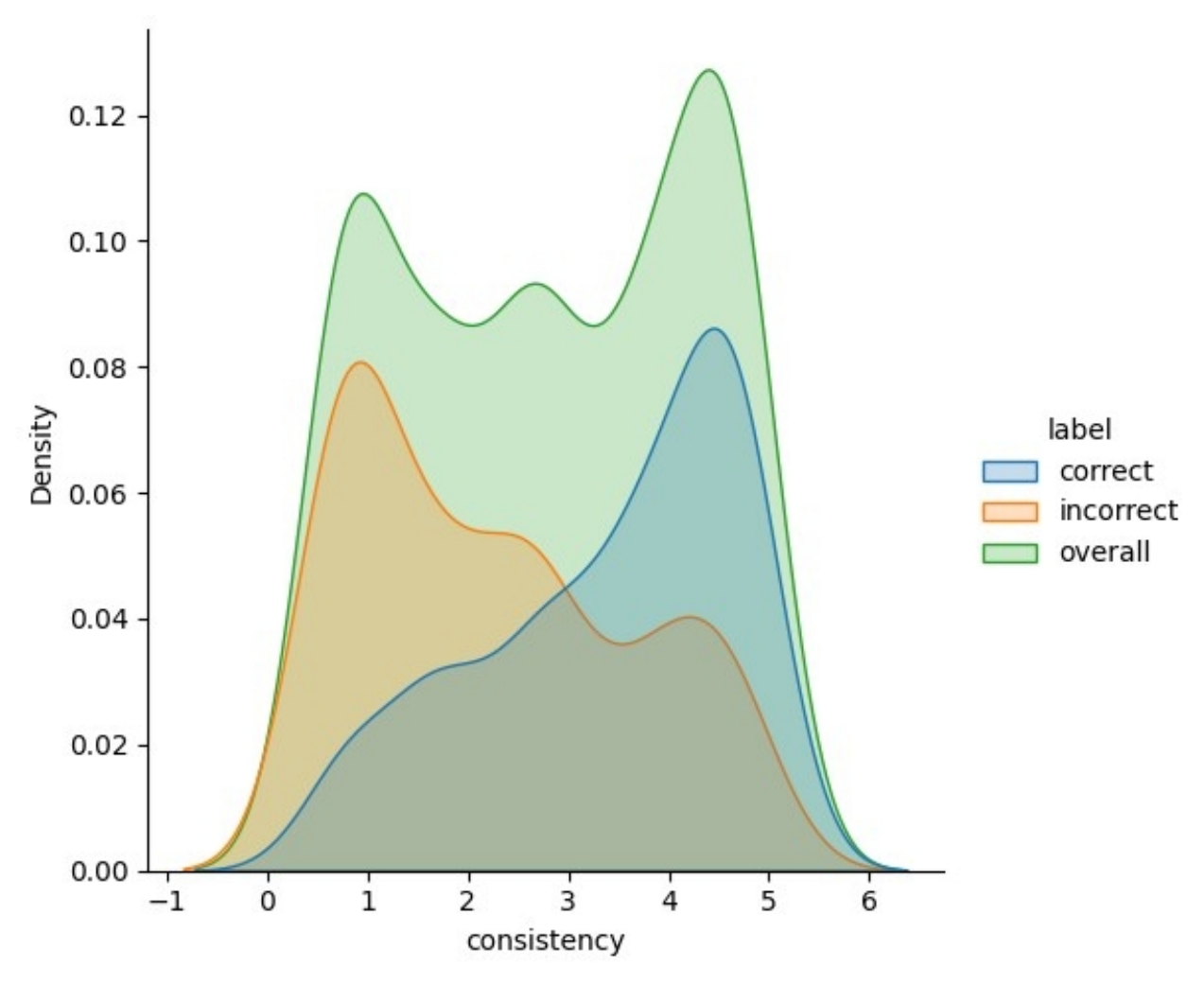}
	}
        \subfigure[Results in \ours+\ve.]{
		\includegraphics[width=0.46\linewidth]{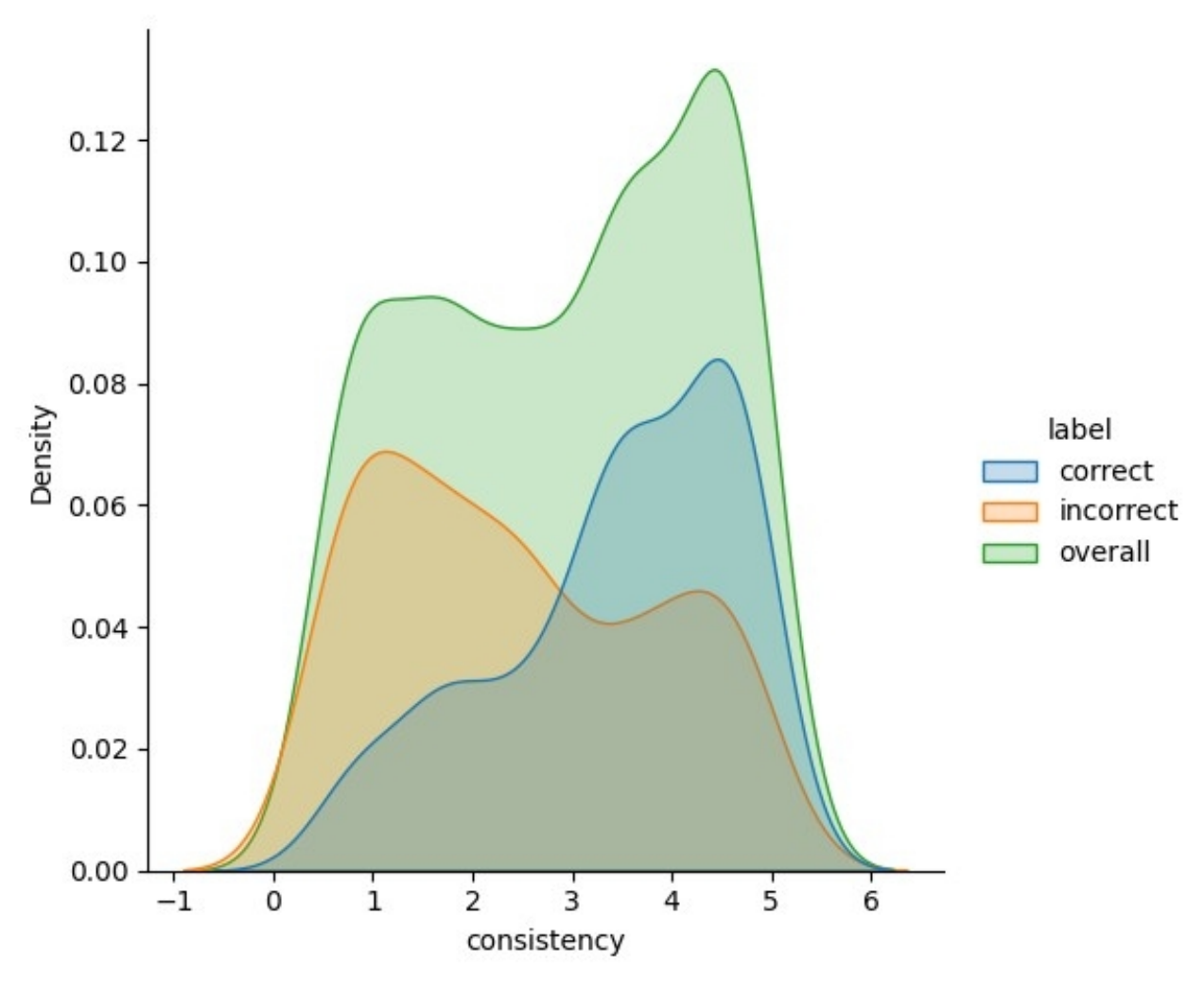}
	}
	\vspace{-0.5ex}
	\caption{Sketch of the consistency distribution in different methods. Consistency denotes the models confidence on predictions, and density denotes the probability density. We present the results in \advhot for illustration. Details in Section~\ref{sec:consistency}. }
    \label{fig:cons}
    \vspace{-3ex}
\end{figure}

\subsection{Consistency Analysis}
\label{sec:consistency}
To explore the confidence of \ours in its determination, we use kernel density estimation~\citep{DBLP:conf/iclr/0002WSLCNCZ23} to analyze the confidence distribution of different methods in generating outputs. 
The results, visualized in Figure~\ref{fig:cons}, show distinct distribution patterns for each method.

From a global perspective, \cotsc, \react, and \ve present a bimodal distribution, while \ours displays a right-skewed distribution.
The distribution illustrates that \ours instills greater confidence in its determinations, which might be derived from the \mii role and the \miii role.
More specifically, \ours exhibits a predominantly right-skewed distribution, with the highest peak value in correct samples compared to others. 
In incorrect samples, however, the distribution presents a bimodal feature, indicating that \ours still make wrong prediction with higher confidence.
The situation is mitigated when integrated with \react and \ve, and we leave the further improvement in this direction as the future work.

\section{Conclusion}
\label{sec:con}

In this paper, we propose \ours a novel and comprehensive approach to enhance reasoning abilities of \llm via acquiring self-denial.
By identifying the challenges, our method builds on the idea of human psyche structure and introduces three tailored roles.
The \mi role tries to provide initial attempts based on LLMs.
Subsequently, the \mii role aims to increase the precision of judgement via summarized rules.
The key points of a specific question are generated as clues for the next phase.
Finally, the \miii role focuses on executable scripts to guide the final refinement, wherein it synthesizes attempts and key points to complete detailed execution.
Besides, we merge three roles into an integration by proposing two-stage fine-tuning strategy to evaluate the resource-friendliness of \ours.
Comprehensive experiments demonstrate that \ours and \ourssft outperform competing models in all primary category, and can be fused to retrieve systems for further enhancement in reasoning accuracy.

\vspace{-0.5ex}

\begin{acks}
This work was partially supported by NSFC (Nos. U23A20296, 72301284, 62272469), and
The Science and Technology Innovation Program of Hunan Province (No. 2023RC1007).
\end{acks}

\bibliographystyle{ACM-Reference-Format}
\bibliography{sample-base}

\appendix

\begin{figure}[!h]
    \centering
    \includegraphics[scale=0.57]{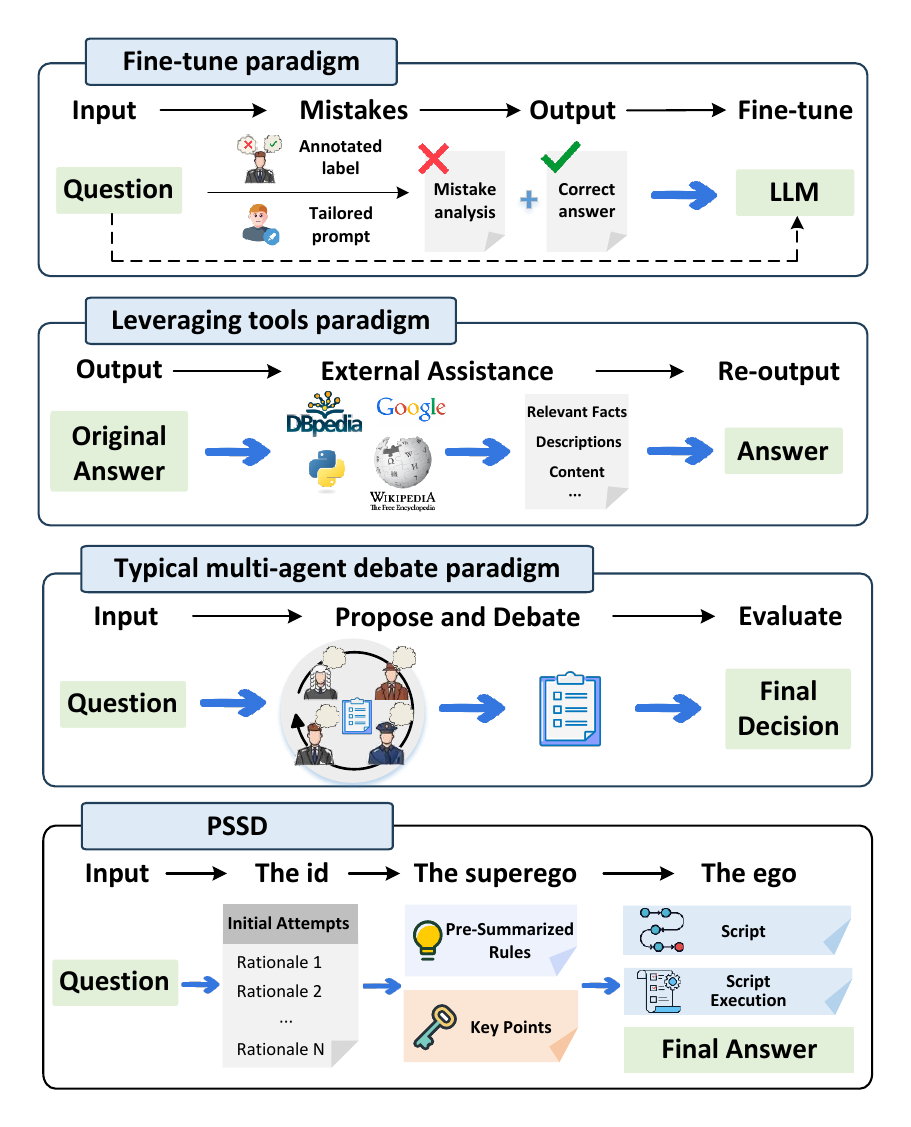}
    \caption{The comparison of different paradigms.}
    \label{fig:comparison}	
\end{figure}

\section{Complementary Experiments}

\subsection{Resource Analysis}
\begin{table}[!h]
\caption{The configuration and the average number of API/LLM calls of multi-agent debate methods. }
\centering
\small
\tabcolsep=0.023\linewidth
\begin{tabular}{lccc}
\hline
\textbf{Method} &  \textbf{\# Agent Number} &  \textbf{\# Debate Round} & \textbf{\# Call} \\ \hline
 \scontrast   &  4   & 1.0 &   7.8     \\
 LM vs LM  &  2  &   2.9   &   8.9      \\
MAD   &   3 & 1.2     & 4.6       \\
Multiagent Debate & 3 &  2.0    &   6.0   \\ 
\hline
\ours &  3 &  1.0    &   5.0 \\  
\ourssft &  1 &  1.0    &   2.0 \\
\hline
\end{tabular}

\label{tab: resource}
\end{table}

\subsection{\ours vs. Self-Contrast}
\label{vs}

\begin{figure}[!h]
	\centering
	\includegraphics[width=1\linewidth]{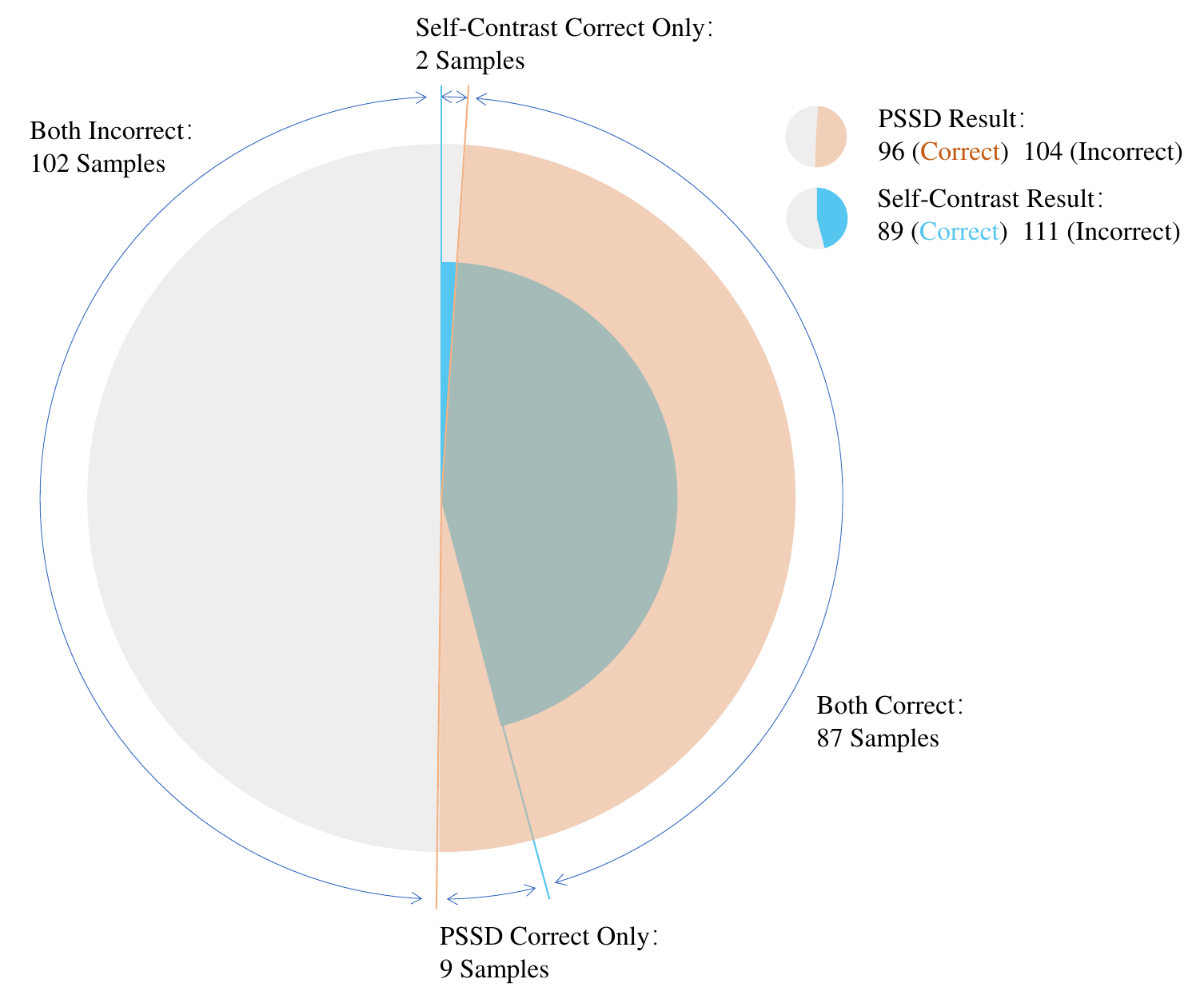}
	\caption{We compare the results of the \ours and \scontrast using two pie charts. It shows \ours is more accurate and stable than direct \scontrast.}
	\label{fig:vs}
\end{figure}

\ours inspires LLMs to engage in human psyche by adopting a multi-agent debate paradigm that is based on the distinct roles of \mi, \mii, and \miii, rather than simply stacking numerous roles or iterating. 
The underlying assumption is that \ours surpasses \scontrast in terms of both accuracy and stability.
To validate this assumption, we conduct an experiment using 200 samples from the \advhot dataset, comparing the results of \ours and \scontrast on each sample. 
As illustrated in Figure~\ref{fig:vs}, \ours achieves higher accuracy with 96 correct answers, compared to 89 correct answers for \scontrast.

We further categorize the results into four distinct cases: (1) Both methods correctly answer; (2) \ours correctly answers, while \scontrast fails; (3) \ours fails, while \scontrast correctly answers; (4) Both methods fail.
The results show that when \scontrast arrives at a correct solution, \ours typically achieves the same outcome, except for two instances where \scontrast succeeded while \ours did not.
However, in 9 instances, \ours succeeded where \scontrast failed, further highlighting \ours's reliability.
These results underscore that \ours not only improves accuracy but also enhances stability, making it a more reliable framework for LLMs reasoning compared to \scontrast.

\subsection{Answering Ability Analysis}
\label{sec:ans}

\begin{figure}[!h]
	\centering
	\includegraphics[width=1\linewidth]{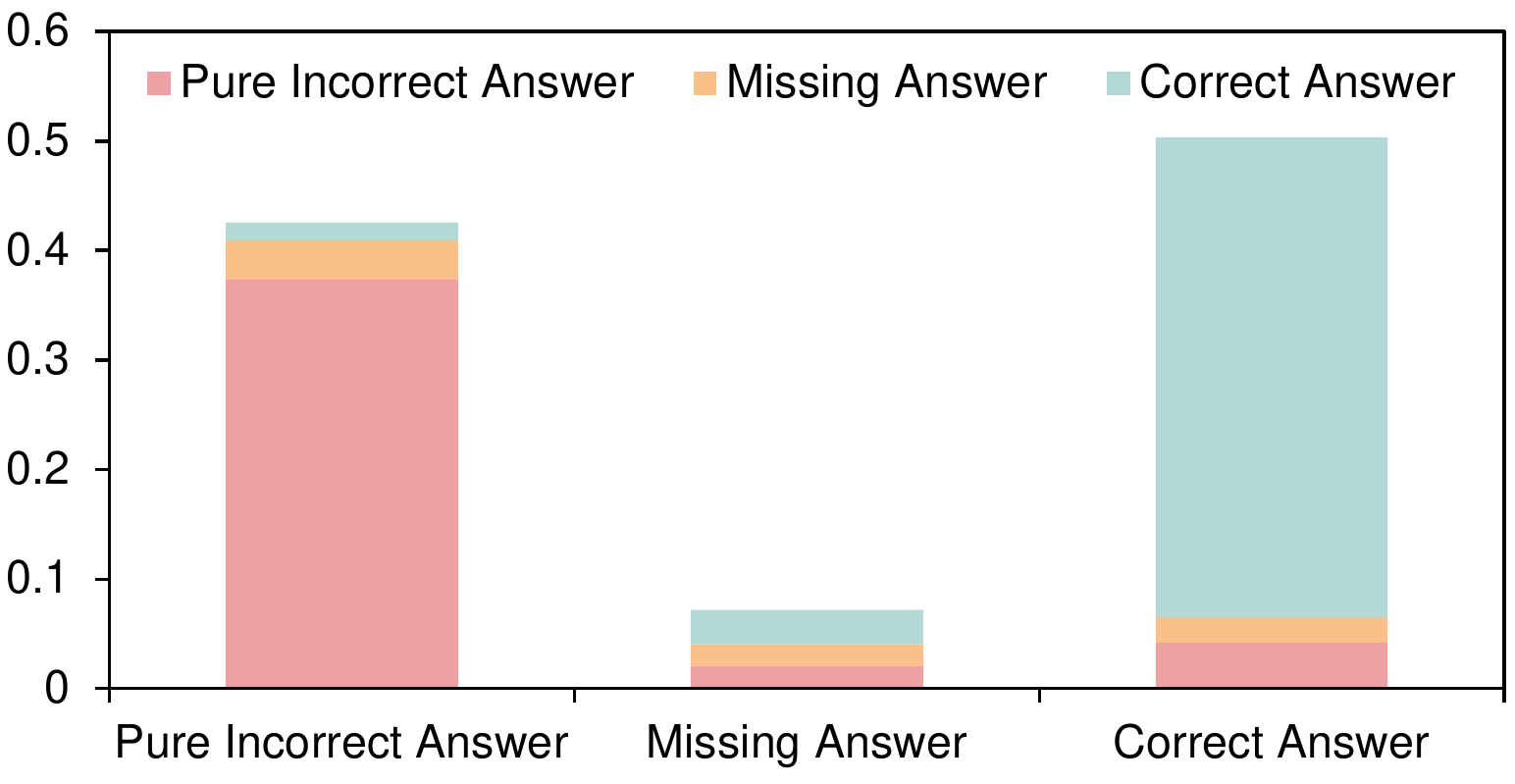}
	\caption{Sketch of the answering type distribution between \cotsc and \ours in \advhot. The presence of the missing answer indicates incorrect results including exact answers, thereby contributing to the overall count of incorrect answers. The height of each bar means the corresponding answer type in \cotsc, and the inside detailed types come from the results of \ours. Details in Section~\ref{sec:ans}.}
	\label{fig:ans}
\end{figure}

The focus lies in harnessing the potential of \llm through learning from mistakes, while it remains crucial to accurately discern the original questions that can be correctly answered. 
Hence, extensive experiments are conducted to closely observe the situation, as depicted in Figure~\ref{fig:ans}.

Evidently, \ours consistently achieves accurate predictions across all answer types and maintains a high level of consistency to the original correct determination (as shown in the right bar). 
For the questions in missing answer type, \ours provides more correct predictions than the incorrect ones while minimizing the number of unresolved cases. This indicates that the proposed method enables \llm to reflect on their generations and carefully select the most satisfactory one. And we leave the efforts on increasing correct hits to the future work. 
Additionally, the presence of three distinct types of outcomes within the original pure incorrect answer type demonstrates that \ours effectively enhances \llm by encouraging more advantageous attempts.

\section{Detailed Statistics for Data}
\label{sec:data}

\begin{table}[!h]
\caption{Detailed statistics for the training and test splits of the datasets we used. 
\textbf{\textsc{Adv.}} means \advhot dataset and \textbf{\textsc{Wiki.}} means \wiki dataset.
 \textbf{\# Key Points} represent the total number of key points contained in each question, while the Average represents the average number of key points per question (calculated in the same way for both \textbf{Scripts} and \textbf{Script Executions}).}
\centering
\footnotesize
\tabcolsep=0.03\linewidth
\begin{tabular}{lcccc}
\toprule
\textbf{Dataset} & \textbf{\textsc{Adv.}}  & \textbf{\textsc{Wiki.}} & \textbf{\gsm} & \textbf{\mat}\\
\cmidrule(){1-1} \cmidrule(l){2-3} \cmidrule(l){4-5}
\begin{tabular}[c]{@{}l@{}}\textbf{Size}\\ \quad- \textit{Training}\\ \quad- \textit{Test} \end{tabular} 
& \begin{tabular}[c]{@{}c@{}}2,620\\ 2,312\\ 308 \end{tabular}
& \begin{tabular}[c]{@{}c@{}}2,554\\ 2,218\\ 336 \end{tabular}
& \begin{tabular}[c]{@{}c@{}}3,087\\ 2,587\\ 500 \end{tabular}
& \begin{tabular}[c]{@{}c@{}}3,092\\ 2,592\\ 500 \end{tabular}\\
\cmidrule(){1-1} \cmidrule(l){2-3} \cmidrule(l){4-5}
\begin{tabular}[c]{@{}l@{}}\textbf{\# Key Points}\\ \quad- \textit{Avg.}\\ \end{tabular}
& \begin{tabular}[c]{@{}c@{}}7,781\\ 2.97\\ \end{tabular}
& \begin{tabular}[c]{@{}c@{}}7,636\\ 2.99\\ \end{tabular}
& \begin{tabular}[c]{@{}c@{}}9,631\\ 3.12\\ \end{tabular}
& \begin{tabular}[c]{@{}c@{}}10,389\\ 3.36\\ \end{tabular}\\
\cmidrule(){1-1} \cmidrule(l){2-3} \cmidrule(l){4-5}
\begin{tabular}[c]{@{}l@{}}\textbf{\# Scripts}\\ \quad- \textit{Avg.}\end{tabular}
& \begin{tabular}[c]{@{}c@{}}8,124\\ 3.10\end{tabular}
& \begin{tabular}[c]{@{}c@{}}7,946\\ 3.11\end{tabular}
& \begin{tabular}[c]{@{}c@{}}12,903\\ 4.18\end{tabular}
& \begin{tabular}[c]{@{}c@{}}13,488\\ 4.36\end{tabular}\\
\cmidrule(){1-1} \cmidrule(l){2-3} \cmidrule(l){4-5}
\begin{tabular}[c]{@{}l@{}}\textbf{\# Script Executions}\\ \quad- \textit{Avg.}\end{tabular}
& \begin{tabular}[c]{@{}c@{}}8,118\\ 3.10\end{tabular}
& \begin{tabular}[c]{@{}c@{}}7,932\\ 3.11\end{tabular}
& \begin{tabular}[c]{@{}c@{}}10,681\\ 3.46\end{tabular}
& \begin{tabular}[c]{@{}c@{}}11,251\\ 3.64\end{tabular}\\
\bottomrule
\end{tabular}

\label{tab:dataset}
\end{table}

\end{document}